\documentclass[10pt,twocolumn,letterpaper]{article}

\usepackage{iccv}
\usepackage{times}
\usepackage{epsfig}
\usepackage{graphicx}
\usepackage{amsmath}
\usepackage{amssymb}
\usepackage{multirow} 
\usepackage{booktabs} 

\newcommand*{\affaddr}[1]{#1} 
\newcommand*{\affmark}[1][*]{\textsuperscript{#1}}
\newcommand*{\email}[1]{\texttt{#1}}

\newcommand{\beginsupplement}{%
        \setcounter{table}{0}
        \renewcommand{\thetable}{S\arabic{table}}%
        \setcounter{figure}{0}
        \renewcommand{\thefigure}{S\arabic{figure}}%
        \setcounter{section}{0}
        \renewcommand{\thesection}{S\arabic{section}}%
     }
     
\usepackage[pagebackref=true,breaklinks=true,letterpaper=true,colorlinks,bookmarks=false]{hyperref}

\newcommand{\minisection}[1]{\vspace{0.04in} \noindent {\bf #1}\ \ }

\iccvfinalcopy 

\begin{document}

\title{Learning the Model Update for Siamese Trackers}

\author{%
Lichao Zhang\affmark[1], Abel Gonzalez-Garcia\affmark[1], Joost van de Weijer\affmark[1], Martin Danelljan\affmark[2], Fahad Shahbaz Khan\affmark[3,4]\\
\affaddr{\affmark[1] Computer Vision Center, Universitat Autonoma de Barcelona, Spain}\\
\affaddr{\affmark[2] Computer Vision Laboratory, ETH Z\"urich, Switzerland}\\
\affaddr{\affmark[3] Inception Institute of Artificial Intelligence, UAE}\\
\affaddr{\affmark[4] Computer Vision Laboratory, Link\"oping University, Sweden}\\
\email{\tt\small \{lichao,agonzalez,joost\}@cvc.uab.es, martin.danelljan@vision.ee.ethz.ch, fahad.khan@liu.se}\\
}

\maketitle

\begin{abstract}
   Siamese approaches address the visual tracking problem by extracting an appearance template from the current frame, which is used to localize the target in the next frame. In general, this template is linearly combined with the accumulated template from the previous frame, resulting in an exponential decay of information over time. While such an approach to updating has led to improved results, its simplicity limits the potential gain likely to be obtained by learning to update. 
   Therefore, we propose to replace the handcrafted update function with a method which learns to update. We use a convolutional neural network, called UpdateNet, which given the initial template, the accumulated template and the template of the current frame aims to estimate the optimal template for the next frame. The UpdateNet is compact and can easily be integrated into existing Siamese trackers. We demonstrate the generality of the proposed approach by applying it to two Siamese trackers, SiamFC and DaSiamRPN.
   Extensive experiments on VOT2016, VOT2018, LaSOT, and TrackingNet datasets demonstrate that our UpdateNet effectively predicts the new target template, outperforming the standard linear update. On the large-scale TrackingNet dataset, our UpdateNet improves the results of DaSiamRPN with an absolute gain of 3.9\% in terms of success score. Code and models are available at \url{https://github.com/zhanglichao/updatenet}.
\end{abstract}

\section{Introduction}

Generic visual object tracking is the task of predicting the location of a target object in every frame of a video, given its initial location.
Tracking is one of the fundamental problems in computer vision, spanning a wide range of applications including video understanding~\cite{renoust2016visual}, surveillance~\cite{emami2012role}, and robotics~\cite{liu2012hand}.
It is a highly challenging task due to frequent appearance changes, various types of occlusions, the presence of distractor objects, and environmental aspects such as motion blur or illumination changes.

\begin{figure}[t]
\centering
    \includegraphics[width=\columnwidth]{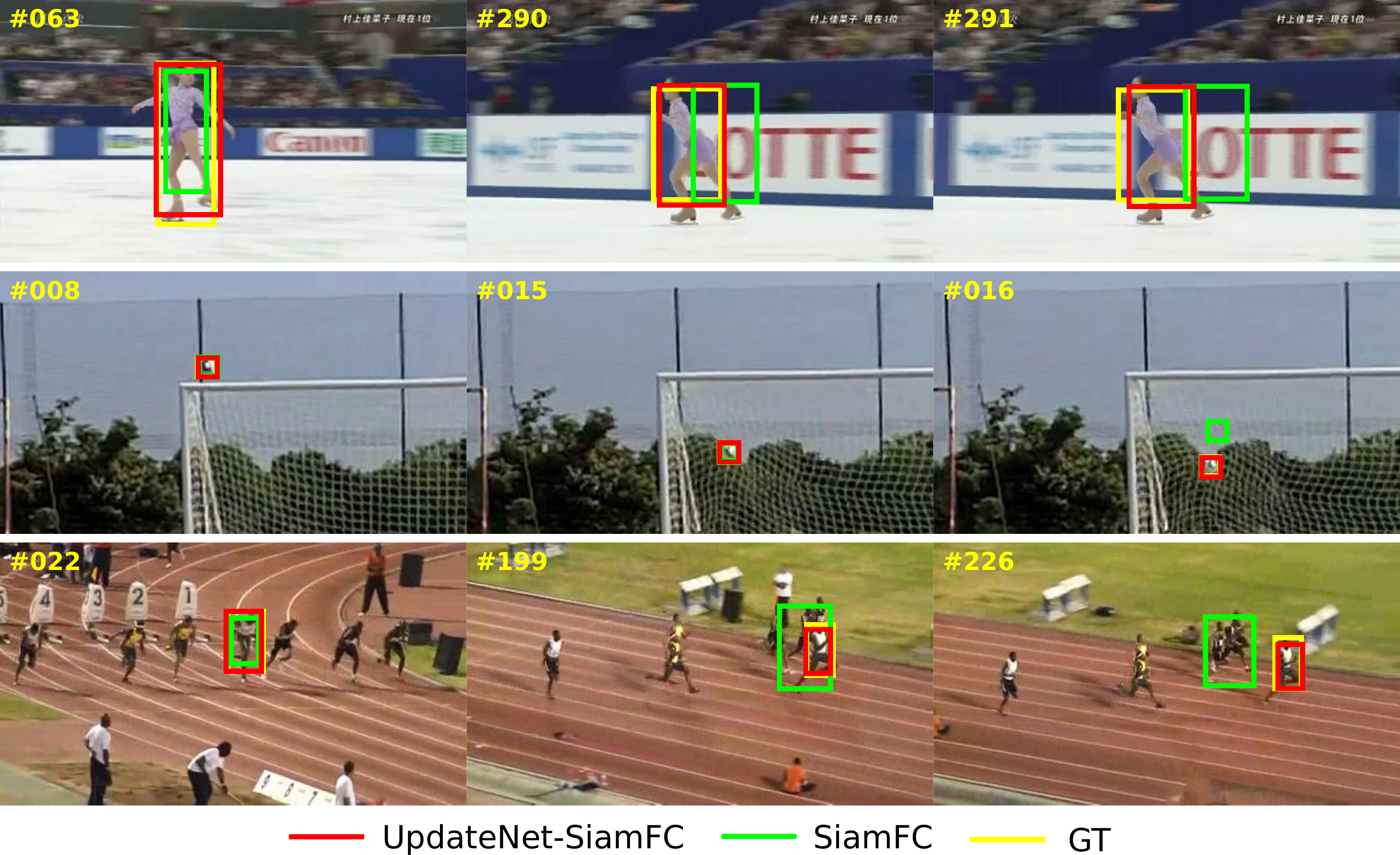}
    \caption{\textbf{Qualitative comparison between model updates.} We learn to update the model template using \emph{UpdateNet}. When combined with Siamese trackers such as SiamFC~\cite{bertinetto2016fully}, our learned updating strategy can be effectively adapted to current circumstances, unlike the simple linear update commonly used. \vspace{-3mm}} 
    \label{fig:intro}
\end{figure}

Currently, there are two prevalent tracking paradigms: Siamese tracking methods~\cite{bertinetto2016fully,li2018high,wang2018learning,zhu2018distractor} and tracking-by-detection methods~\cite{bolme2010visual,danelljan2017eco,danelljan2016beyond,henriques2015high,nam2016learning,zhang2014fast,zhang2019learning}.
In this work, we consider Siamese trackers, since they provide competitive accuracy while achieving impressive computational efficiency.
The basic principle of these trackers is to match an object appearance template with a corresponding feature representation of the search region in the test frame.
The features for the object template and the search region are acquired through a deep neural network trained offline on a large dataset. Such a training strategy has shown to provide excellent visual descriptors for the tracking task \cite{bertinetto2016fully,zhu2018distractor}.

In the original Siamese tracker~\cite{bertinetto2016fully}, the object template is initialized in the first frame and then kept fixed during the remainder of the video. However, appearance changes are often large and failing to update the template could lead to early failure of the tracker. In such scenarios, it is important to adapt the model to the current target appearance. To accommodate this problem, more recent Siamese trackers~\cite{li2018high,wang2018learning,zhu2018distractor} have implemented a simple linear update strategy using a running average with a fixed learning rate~\cite{stauffer1999adaptive}. 
This strategy assume a constant rate of appearance change across all frames in the video, as well as across different videos. 
In practice, the update requirements for the object template greatly vary for different tracking situations, which depend on a complex combination of external factors such as motion, blur, or background clutter.
Therefore, a simple linear update is often inadequate to cope with changing updating needs and to generalize to all potentially encountered situations.
Moreover, this update is also constant in all spatial dimensions, which does not allow for localized partial updates.
This is especially damaging in situations such as partial occlusions, where only a certain part of the template needs to be updated.
Finally, excessive reliance on the initial template may suffer from catastrophic drift and the inability to recover from tracking failures.

In this paper, we propose to \emph{learn} the target template update itself. Our learned update strategy utilizes target and image information, and is thus adaptive to the present circumstances of each particular situation.
In our approach, the updated template is computed as a function of (i) the initial ground-truth template, (ii) the accumulated template from all previous frames, and (iii) the feature template at the predicted object location in the current frame.
Hence, the new accumulated template contains an effective historical summary of the object's current appearance, as it is continually updated using the most recent information while being robust due the strong signal given by the initial object appearance.
More specifically, the aforementioned template update function is implemented as a convolutional neural network, \emph{UpdateNet}.
This is a compact model that can be combined with any Siamese tracker to enhance its online updating capabilities while maintaining its efficiency properties.
Furthermore, it is sufficiently complex to learn the nuances of effective template updating and be adaptive enough to handle a large collection of tracking situations.

We evaluate UpdateNet by combining it with two state-of-the-art Siamese trackers: SiamFC~\cite{bertinetto2016fully} and DaSiamRPN~\cite{zhu2018distractor}.
Through extensive experiments on common tracking benchmarks, such as VOT2018~\cite{kristan2018sixth}, we demonstrate how our UpdateNet provides enhanced updating capabilities that in turn result in improved tracking performance (see Figure~\ref{fig:intro}).   
We also present results in the recent LaSOT dataset~\cite{fan2018lasot}, which is substantially more challenging as it contains abundant long-term sequences. 
Overall, we propose an efficient model to learn how to effectively update the object template during online tracking and that can be applied to different existing Siamese trackers.

\vspace{-1mm}
\section{Related work}
\vspace{-2mm}
\minisection{Tracking Frameworks.}
Most existing tracking methods are either based on tracking-by-detection or employ template matching. Object trackers based on tracking-by-detection pose the task of target localization
as a classification problem where the decision boundary is obtained by online learning a discriminative classifier using image patches from the target object and the background. Among the tracking-by-detection approaches, discriminative correlation filter based trackers~\cite{henriques2015high,zhang2019learning,danelljan2016beyond,danelljan2017eco} have recently shown excellent performance on several tracking benchmarks~\cite{wu2013otb,wu2015object,kristan2016VOT,kristan2018sixth}. These trackers learn a
correlation filter from example patches of the target appearance to discriminate between the target and background
appearance. 

The other main tracking framework is based on template matching, typically using Siamese networks~\cite{bertinetto2016fully,valmadre2017end,wang2018learning,held2016learning,li2018high,zhu2018distractor}, that implement a similarity network by spatial cross-correlation. Bertinetto \etal~\cite{bertinetto2016fully} proposed a Siamese tracker which is based on a two-stream architecture. One stream extracts the object template's features based on an \emph{exemplar} image that contains the object to be tracked. 
The other stream receives as input a large \emph{search region} in the target image. 
The two outputs are cross-correlated to generate a response map of the search region. Many trackers have extended the SiamFC architecture~\cite{valmadre2017end,he2018twofold,wang2018learning,li2018high,zhu2018distractor,zhang2018structured} for tracking. The Siamese-based trackers have gained popularity since they provide a good trade-off between computational speed and tracking performance. However, most of these approaches struggle to robustly classify the target especially in the presence of distractors due to no online learning. In this work, we analyze the limitations of Siamese trackers regarding the update of the template model and propose a solution to address them.

\minisection{Updating the object template}
Most trackers either use simple linear interpolation to update the template in every frame~\cite{bolme2010visual,henriques2015high,danelljan2015learning,danelljan2016beyond,kiani2017learning,choi2018context} or do not update the initial template at all~\cite{bertinetto2016fully,wang2018learning,li2018high,zhu2018distractor}.
Such update mechanisms are insufficient in most tracking situations, as the target object may suffer appearance changes given by deformations, fast motion, or occlusion.
Moreover, fixed update schedules also result in object templates that are more focused on recent frames~\cite{danelljan2016adaptive}, while forgetting the historical appearances of the object.
To address this issue, Danelljan \etal~\cite{danelljan2016adaptive,danelljan2016beyond} propose to include a subset of historic frames as training samples when calculating the current correlation filter, which leads to better results than the traditional linear frame-by-frame update. 
Nonetheless, storing multiple samples in memory results in increased computation and memory usage, which in turn heavily decreases the tracking speed. 
The ECO tracker~\cite{danelljan2017eco} tries to alleviate this problem by modelling the distribution of training samples as a mixture of Gaussians, where each component represents a distinct appearance.
This significantly reduces required storage and, combined with a conservative update strategy (only every five frames), leads to increased tracking efficiency.
Even with more previous samples, the correlation filter is still updated by averaging the filters of their corresponding samples (as still linear interpolation update).

Recently, 
Yang \etal~\cite{yang2018learning} employed a Long Short-Term Memory (LSTM) to estimate the current template by storing previous templates in memory during on-line tracking, which is computationally expensive and a rather complex system.
Choi \etal~\cite{choi2017visual} also uses a template memory but uses reinforcement learning to select one of the stored templates. This method fails to accumulate information from multiple frames.
The meta-tracker of~\cite{park2018meta} extends the initialization of the target model in the first frame by a pre-trained approach, but still need a linear update in on-line tracking.
Yao \etal~\cite{yao2018joint} propose to learn the updating coefficients for CF trackers using SGD offline. 
While the solution for correlation filter is still the hand-crafted manner, and these coefficients are fixed without updating during tracking.

To adapt to the object variations, Guo \etal~\cite{guo2017learning} propose to compute a transformation matrix with respect to the initial template through regularized linear regression in the Fourier domain.
Since only the initial template is considered when estimating the transformation, this approach ignores the historical object variations observed during tracking, which may result important for a smoother adaptation of the exemplar template.
Moreover, they compute the transformation matrix as a closed-form solution on the Fourier domain, which suffers from issues related to the boundary effect~\cite{kiani2015correlation}.
Our work instead uses a powerful yet easily trainable model to update the object template based not only the first frame, but also on the accumulated template using all previous frames, leveraging the observed training data. 
Furthermore, our UpdateNet is trained to \emph{learn} how to perform an effectively update the object template, based on the observed training tracking data.

\begin{figure*}[t!]
    \centering
    \includegraphics[width=\textwidth]{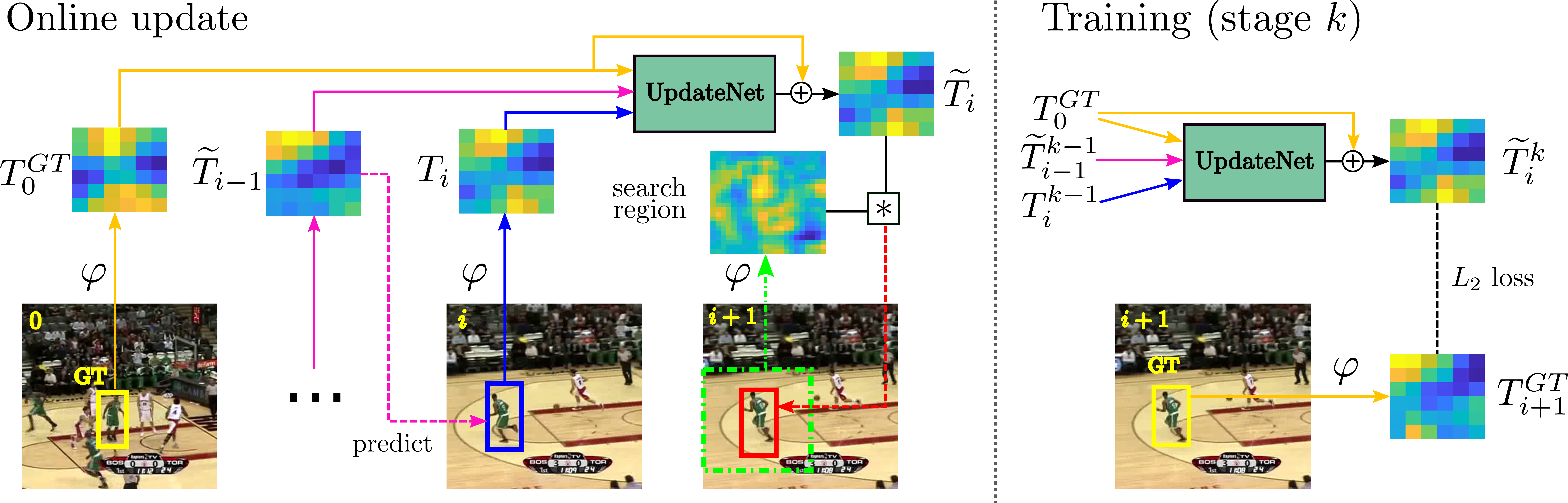}
    \caption{\textbf{Overview of our tracking framework with UpdateNet.} (Left) The online update of the object template is performed by UpdateNet, which receives as input the initial ground-truth template, last accumulated template and current predicted template, and outputs updated accumulated template. (Right) Training of UpdateNet using the distance to the ground-truth object template on next frame.}
    \vspace{-4mm}
    \label{fig:pipeline}
\end{figure*}

\vspace{-1mm}
\section{Updating the object template}
\vspace{-1mm}
In this section, we present our approach to learn how to update the object template during online tracking. 
We start by revisiting the standard update mechanism in tracking and identifying its drawbacks.
Then, we introduce our formulation to overcome them and describe our model and training procedure in detail. 
The focus of this paper is on Siamese trackers.
Note, however, that our approach is not limited to Siamese trackers and the same formulation could be applied to other types of trackers, e.g.\ DCF~\cite{henriques2015high,danelljan2016beyond,danelljan2017eco}.
\vspace{-1mm}
\subsection{Standard update}
\vspace{-1mm}
Several recent tracking approaches~\cite{bolme2009average,bolme2010visual,henriques2015high,valmadre2017end,wang2018learning,li2018high,zhu2018distractor} use a simple averaging strategy to update the object appearance model given a new data sample. 
This strategy dates from early tracking methods~\cite{stauffer1999adaptive} and has long been the standard for online updating given its acceptable results and in spite of its limitations.
The template is updated as a running average with exponentially decaying weights over time. The choice of exponential weights yields the following recursive formula for updating the template,
\begin{align}
\vspace{-1mm}
  \widetilde{T}_i = (1-\gamma)\widetilde{T}_{i-1} + \gamma T_i.
  \vspace{-2mm}
  \label{eq:linear_update}
\end{align}
Here, $i$ is the frame index, $T_i$ is the new template sample computed using only the current frame, and $\widetilde{T}_i$ is the accumulated template.
The update rate $\gamma$ is normally set to a fixed small value (e.g.\ $\gamma=0.01)$ following the assumption that the object's appearance changes smoothly and consistently in consecutive frames.
In DCF trackers (e.g.~\cite{bolme2010visual,henriques2015high}), $T$ corresponds to the correlation filter.
In Siamese trackers, instead, $T$ is the object appearance template extracted from a particular frame by a fully convolutional feature extractor. 
While the original SiamFC tracker~\cite{bertinetto2016fully} does not perform any model update, more recent Siamese trackers~\cite{bertinetto2016fully,wang2018learning,li2018high,zhu2018distractor} have adopted~\eqref{eq:linear_update} to update their templates.

While template averaging provides a simple means of integrating new information, it has several severe drawbacks:
\begin{itemize}
    \vspace{-2mm}
    \item It applies a constant update rate for every video, despite the possibly different updating necessities caused by multiple factors such as camera motion. Even within the same video, the required update on the object template may dynamically vary at different times.\vspace{-2.5mm}
     \item The update is also constant along all spatial dimensions of the template, including the channel dimension. This prevents updating only part of the template, which is desirable under partial occlusions, for example.  \vspace{-2.5mm} 
     \item The tracker cannot recover from drift. Partially, this is caused by the fact that it loses access to the appearance template $T_0$, which is the only template which is without doubt on the object. 
     \vspace{-2.5mm}
     \item The update function is constrained to a very simple linear combination of previous appearance templates. This severely limits the flexibility of the update mechanism, important when the target undergoes complex appearance changed. Considering more complex combination functions is expected to improve results. 
\end{itemize}

\vspace{-3mm}
\subsection{Learning to update}
\vspace{-1mm}
We address the drawbacks listed above by proposing a model that learns an adaptive update strategy.
Since the focus of this paper is on Siamese trackers, $T$ represents here the object appearance template.  
To address the limitations of simple template averaging, we propose to learn a generic function $\phi$ that updates the template according to,
\begin{align}
\vspace{-1mm}
  \widetilde{T}_i = \phi (T_0^{GT}, \widetilde{T}_{i-1}, T_i).
  \vspace{-3mm}
  \label{updatenet}
\end{align}
The learned function $\phi$ computes the updated template based on initial ground-truth template $T_0^{GT}$, the last accumulated template $\widetilde{T}_{i-1}$ and the template $T_i$ extracted from the predicted target location in the current frame. In essence, the function updates the previous accumulated template $\widetilde{T}_{i-1}$ by integrating the new information given by the current frame $T_i$.
Therefore, $\phi$ can be adapted to the specific updating requirements of the current frame, based on the difference between the current and accumulated templates.
Moreover, it also considers the initial template $T_0^{GT}$ in every frame, which provides highly reliable information and increases robustness against model drift.
The function $\phi$ is implemented as a convolutional neural network, which grants great expressive power and the ability to learn from large amounts of data.
We call this neural network \emph{UpdateNet} and describe it in detail in the following section. 
\vspace{-1mm}
\subsection{Tracking framework with UpdateNet}
\vspace{-1mm}
We present here the structure of UpdateNet and describe how it is applied for online tracking. Figure~\ref{fig:pipeline} (left) presents an overview of our adaptive object update strategy using UpdateNet with a Siamese tracker. 
We extract deep features from image regions with a fixed fully convolutional network $\varphi$, employing the same feature extractor as in the SiamFC tracker~\cite{bertinetto2016fully}.
We extract $T_0^{GT}$ from the ground-truth object location in the initial frame (number 0 in Figure~\ref{fig:pipeline}).
In order to obtain $T_i$ for the current frame, we use the accumulated template from all previous frames $\widetilde{T}_{i-1}$ to predict the object location in frame $i$ (dashed purple line) and extract the features from this region (solid blue line).
Note that $\widetilde{T}_{i-1}$ corresponds to the output of UpdateNet for the previous time step, not shown here for conciseness.
We concatenate the extracted features $T_0^{GT}$ and $T_i$ with the accumulated features $\widetilde{T}_{i-1}$ to form the input of UpdateNet. 
This input is then processed through a series of convolutional layers (sec.~\ref{sec:details}) and outputs the predicted new accumulated template $\widetilde{T}_{i}$.
For the first frame, we set $T_i$ and $\widetilde{T}_{i-1}$ to $T_0^{GT}$ as there have not been any previous frames.

The only ground-truth information used by UpdateNet is the given object location in the initial frame, all other inputs are based on predicted locations.
Hence, $T_0^{GT}$ is the most reliable signal on which UpdateNet can depend to guide the update.
For this reason, we employ a residual learning strategy~\cite{he2016deep}, where UpdateNet learns how to modify the ground-truth template $T_0^{GT}$ for the current frame. This is implemented by adding a skip connection from $T_0^{GT}$ to the output of UpdateNet.
This approach still takes into account the set of historical appearances of the object for updating, but pivots such update on the most accurate sample. 
We have also experimented with adding skip connections from other inputs as well as no residual learning at all (see sec.~\ref{sec:experiments}).
\vspace{-1mm}
\subsection{Training UpdateNet}
\vspace{-1mm}
\label{sec:trainingUpdateNet}

We train our UpdateNet to predict the target template in the coming frame, i.e.\ the predicted template $\widetilde{T}_i$ should match the template $T_{i+1}^{GT}$ extracted from the ground-truth location in the next frame (Figure~\ref{fig:pipeline}, right). The intuition behind this choice is that $T_{i+1}^{GT}$ is the optimal template to use when searching for the target in the next frame.
In order to achieve this, we train UpdateNet by minimizing the Euclidean distance between the updated template and the ground-truth template of the next frame, defined as
\begin{equation}
    \mathcal{L}_2 = \left \lVert  \phi (T_0^{GT}, \widetilde{T}_{i-1}, T_i) - T_{i+1}^{GT} \right\rVert_2.
    \label{eq:loss}
\end{equation}
In the remainder of this section we describe the procedure employed to generate the training data and introduce a multi-stage training approach for UpdateNet.

\minisection{Training samples.} In order to train UpdateNet to minimize~\eqref{eq:loss}, we need pairs of input triplets $(T_0^{GT}, \widetilde{T}_{i-1}, T_i)$ and outputs $T_{i+1}^{GT}$ that reflect the updating needs of the tracker during online application.
The object templates of the initial frame $T_0^{GT}$ and target frame $T_{i+1}^{GT}$ can be easily obtained by extracting features from ground-truth locations in corresponding frames.
In case of the current frame's template $T_i$, however, using the ground-truth location represents a seldom encountered situation in practice, for which the predicted location in the current frame is very accurate.
This unrealistic assumption biases the update towards expecting very little change with respect to $T_i$, and thus UpdateNet cannot learn a useful updating function.
Therefore, we need to extract $T_i$ samples for training by using an imperfect localization in the $i$-th frame.
We can simulate such situation by using the accumulated template $\widetilde{T}_{i-1}$, ideally presenting localization errors that occur during online tracking.

\minisection{Multi-stage training.} In theory, we could use the accumulated template $\widetilde{T}_{i-1}$ output by UpdateNet.
However, this would force the training to be recurrent, making the procedure cumbersome and inefficient.
To avoid this, we split our training procedure into sequential stages that iteratively refine UpdateNet.
In the first stage, we run the original tracker on the training dataset using the standard linear update
\begin{equation}
\vspace{-1mm}
 \widetilde T_i^0  = \left( {1 - \gamma } \right)\widetilde T_{i - 1}^0  + \gamma T_i^0, 
 \vspace{-1mm}
\end{equation}

which generates accumulated templates and realistically predicted locations for each frame. 
We set the update rate $\gamma$ to the recommended value for the tracker.
This corresponds to a first approximation to likely inputs for UpdateNet during tracking inference, albeit with the less sophisticated linear update strategy.
In every posterior training stage $k\in \{1,...,K\}$, we use the UpdateNet model trained in the previous stage to get accumulated templates and object location predictions as follows
\begin{equation}
 \vspace{-1mm}
    \widetilde T_i^k  = \phi _i^k \left( {T^{GT}_0 ,\widetilde T_{i - 1}^{k - 1} ,T_i^{k - 1} } \right) \,. 
    \vspace{-1mm}
\end{equation}
Such training data samples closely resemble the expected data distribution at inference time, as they have been output by UpdateNet.
We investigate a suitable value for the total number of stages $K$ in the experimental section (sec.~\ref{sec:experiments}).

\section{Experiments}
\label{sec:experiments}
\subsection{Training dataset}

We use the recent Large-scale Single Object Tracking (LaSOT)~\cite{fan2018lasot} to train our UpdateNet. 
LaSOT has 1,400 sequences in 70 categories, which amounts to a total of 3.52M frames. 
Each category contains exactly twenty sequences, making the dataset balanced across classes. 
It also provides longer sequences that contain more than 1,000 frames (2,512 frames on average) in order to satisfy the current long-term trend in tracking. 
We used the official training and test splits, which preserve the balanced class distribution.
In fact, we only employ a subset containing 20 training sequences from 20 randomly selected categories, with a total of 45,578 frames. 
We have found experimentally that this suffices to learn an effective updating strategy, and that additional data brings only a small performance boost while increasing training time.

\subsection{Evaluation datasets and protocols}
We evaluate results on standard tracking benchmarks: VOT2018/16~\cite{VOT_TPAMI}, LaSOT~\cite{fan2018lasot} and TrackingNet~\cite{muller2018trackingnet}.

\minisection{VOT2018/16~\cite{VOT_TPAMI}.}
VOT2018 dataset has 60 public testing sequences, with a total of 21,356 frames. It is used as the most recent edition of the VOT challenge. 
The VOT protocol establishes that when the evaluated tracker fails, i.e.\ when the overlap with the ground-truth is below a given threshold, it is re-initialized in the correct location five frames after the failure. 
The main evaluation measure used to rank the trackers is Expected Average Overlap (EAO), which is a combination of accuracy (A) and robustness (R). 
We also use VOT2016~\cite{kristan2016VOT} for comparison purposes, which has 10 different sequences with VOT2018~\cite{kristan2018sixth}.
We compute all results using the provided toolkit~\cite{kristan2018sixth}.

\minisection{LaSOT~\cite{fan2018lasot}.}
LaSOT is a much larger and more challenging dataset including long-term sequences, 
Following recent works that use this dataset~\cite{li2018siamrpn++,fan2018siamese}, we report results on protocol II, i.e.\ LaSOT testing set. The testing subset has 280 sequences with 690K frames in total. 
LaSOT dataset~\cite{fan2018lasot} follows the OPE criterion of OTB~\cite{wu2013otb}. It consists of precision plot which is measured by the center location error, and success plot which is measured through the intersection over union (IoU) between the predicted bounding box and the ground-truth.
Besides precision plot and success plot, LaSOT also uses normalized precision plot to counter the situation that target size and image resolution have large discrepancies for different frames and videos, which heavily influences the precision metric. We use success plot and normalized precision plot to evaluate the trackers in the article.
We use their code~\cite{fan2018lasot} to create all plots.

\minisection{TrackingNet~\cite{muller2018trackingnet}.}
This is a large-scale tracking dataset consisting of videos in the wild. 
It has a total of 30,643 videos split into 30,132 training videos and 511 testing videos, with an average of 470,9 frames.
It uses precision, normalized precision and success as evaluation metrics.

\vspace{-1mm}
\subsection{Implementation details}
\vspace{-1mm}
\label{sec:details}
We use SiamFC~\cite{bertinetto2016fully} and DaSiamRPN~\cite{zhu2018distractor} as our base trackers, and the backbone Siamese network adopts the modified AlexNet. We do not perform any changes except for in the updating component.
The original implementation of SiamFC did not perform any object update. We borrow a linear update rate from CFNet~\cite{valmadre2017end} as $\gamma=0.0102$ for templates generating in the training stage 1. 
We use the original version of DaSiamRPN, which does not employ any update strategy.
We analyze the effect of the linear update rate in the tracking performance in sec.~\ref{sec:update_date}.
To train UpdateNet, we set a group of templates, including $T_0^{GT}$, $\widetilde{T}_{i-1}$, $T_i$ and $T_{i+1}^{GT}$ as the input. They are all sampled sequentially from a same video. It is noteworthy that $\widetilde{T}_{i-1}$ and $T_i$ are generated by the real tracking procedure, while $T_0^{GT}$ and $T_{i+1}^{GT}$ are ground-truth templates.
We store all training object templates on disk, extracted using either linear/no update (stage $k=1$) or a previous version of UpdateNet ($k>1$).
Let the template size be $H\times W\times C$. 
UpdateNet is a two-layer convolutional neural network: one $1\times 1\times 3\cdot C\times 96$ convolutional layer, followed by a ReLU and a second convolutional layer of dimensions $1\times 1\times 96\times C$.
For SiamFC, $H=W=6$ and $C=256$, whereas DaSiamRPN $C=512$. 
In the first stage, the weights are initialized from scratch and the learning rate is decreased logarithmically at each epoch from $10^{-6}$ to $10^{-7}$. 
In next stage, the weights are initialized by the best model from the last stage, and the learning rate is decreased logarithmically at each epoch from $10^{-7}$ to $10^{-8}$. 
We train the model for 50 epochs with mini-batches of size 64. 
 We use Stochastic Gradient Descent (SGD) with momentum of 0.9 and weight decay of 0.0005.

\begin{table}[t]
    \begin{center}      
        \begin{tabular}{lcccc}
            \toprule
            Update for SiamFC & Skip & EAO ($\uparrow$)	& A ($\uparrow$) & R ($\downarrow$)\\         
            \midrule
Linear  & -   & 0.188 			&0.50	            &0.59 \\
\midrule
 UpdateNet ($K=1$)                  & - 	    & 0.205 			&0.48 		    	&0.58 \\   
 UpdateNet ($K=1$)& $T_i$    	&0.207 			&0.47				&0.57\\ 
 UpdateNet ($K=1$) &  $\widetilde{T}_{i-1}$      & 0.214 			&0.49 				&0.58 \\ 
 UpdateNet ($K=1$) &  $T_0^{GT}$        & 0.250 			&0.50 				&0.53 \\ 
\midrule
 UpdateNet ($K=2$) & $T_0^{GT}$ & 0.257 			&0.51 				&0.50 \\
 UpdateNet ($K=3$) & $T_0^{GT}$   &\textbf{0.262}  			&\textbf{0.52} 				&\textbf{0.49} \\ \bottomrule         
        \end{tabular}
        \vspace{2mm}
        \caption{\small \textbf{Ablation study on VOT2018~\cite{kristan2018sixth}.} We present several update strategies for SiamFC~\cite{bertinetto2016fully}. The results are reported in terms of EAO, normalized weighted mean of accuracy (A), and normalized weighted mean of robustness score (R). `Skip' column indicates the origin of the skip connection, if any. Here, $K$ is the number of stages UpdateNet is trained for.} 
        \label{table:ablation}
        \vspace{-6mm}
    \end{center}
    
\end{table}

\vspace{-1mm}
\subsection{Ablation study}
\vspace{-1mm}
We start our evaluation by ablating our approach on different components in order to analyze their contribution to the final performance. 
Table~\ref{table:ablation} shows the results using VOT2018~\cite{kristan2018sixth} dataset under the EAO measure.
In the middle of the table, it shows that updating the object template with the first stage of UpdateNet results beneficial when it is residually trained with respect to $T_0^{GT}$, as the learned update strategy is grounded on a reliable object sample.
Moreover, our multi-stage training further increases the performance achieved by UpdateNet, reaching a total improvement of 7.4\% with respect to the original SiamFC with no update.
For the remainder of the paper, we use UpdateNet trained on 3 stages and with skip connection from $T_0^{GT}$.

\begin{figure}[!t]
    \centering
    \includegraphics[width=1\columnwidth]{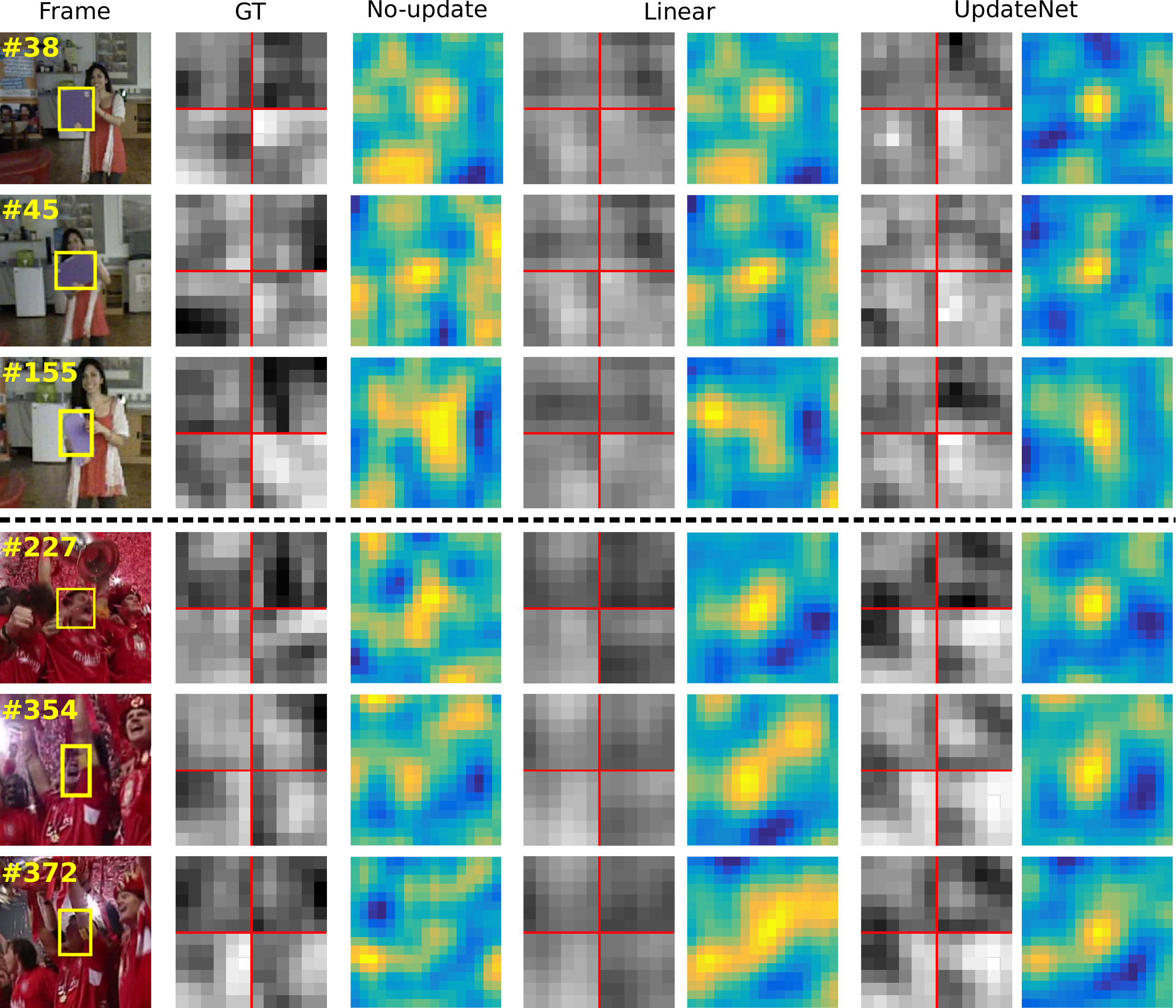}
     \caption{\small \textbf{Visualization of accumulated templates for SiamFC.} `Frame' column shows the search region and ground-truth box used to extract the templates, whose top four channels we show in  `GT'.
     `No-update' presents the response map resulting from applying the inital template to the search region. For `Linear' and `UpdateNet' stategies we also show their accumulated templates. \vspace{-4mm} }
     \label{fig:templates}
\end{figure}

\vspace{-1mm}
\subsection{Analysis on representation update}
\vspace{-1mm}
This section attempts to provide insights regarding the performance improvements achieved by UpdateNet.
Siamese networks are trained to project the images into a feature space in which spatial correlation is maximal.
Update strategies operate on the learned features, possibly interfering with their correlation abilities and potentially damaging the tracking performance.
In order to study the interference of the update strategies on the features we visualize the accumulated templates of SiamFC for both linear update and UpdateNet in Figure~\ref{fig:templates}.
We also include the ground-truth template extracted from the annotated bounding-box.
For each template we show the feature maps of the four most dynamic channels in the ground-truth template, arranged in a 2$\times$2 grid. 
For comparison reasons, the accumulated templates are generated using the ground-truth object locations instead of the predicted locations during tracking.
Moreover, next to each accumulated template we also show the response map generated when correlating the template with the search region.
We observe several interesting properties that support the performance gains seen in practice.
First, the accumulated templates using UpdateNet resemble the ground-truth more closely than those in linear update (see e.g. highlight on bottom-right channel for frame 38, first example).
Second, response maps tend to be sharper on the object location for UpdateNet, which shows how our strategy does not negatively interfere with the desired correlation properties of the learned features.
Finally, the accumulated templates of the linear update change at a very slow rate and are clearly deficient in keeping up with the appearance variation exhibited in videos.

\begin{figure}[!t]
    \centering
    \includegraphics[width=.95\columnwidth]{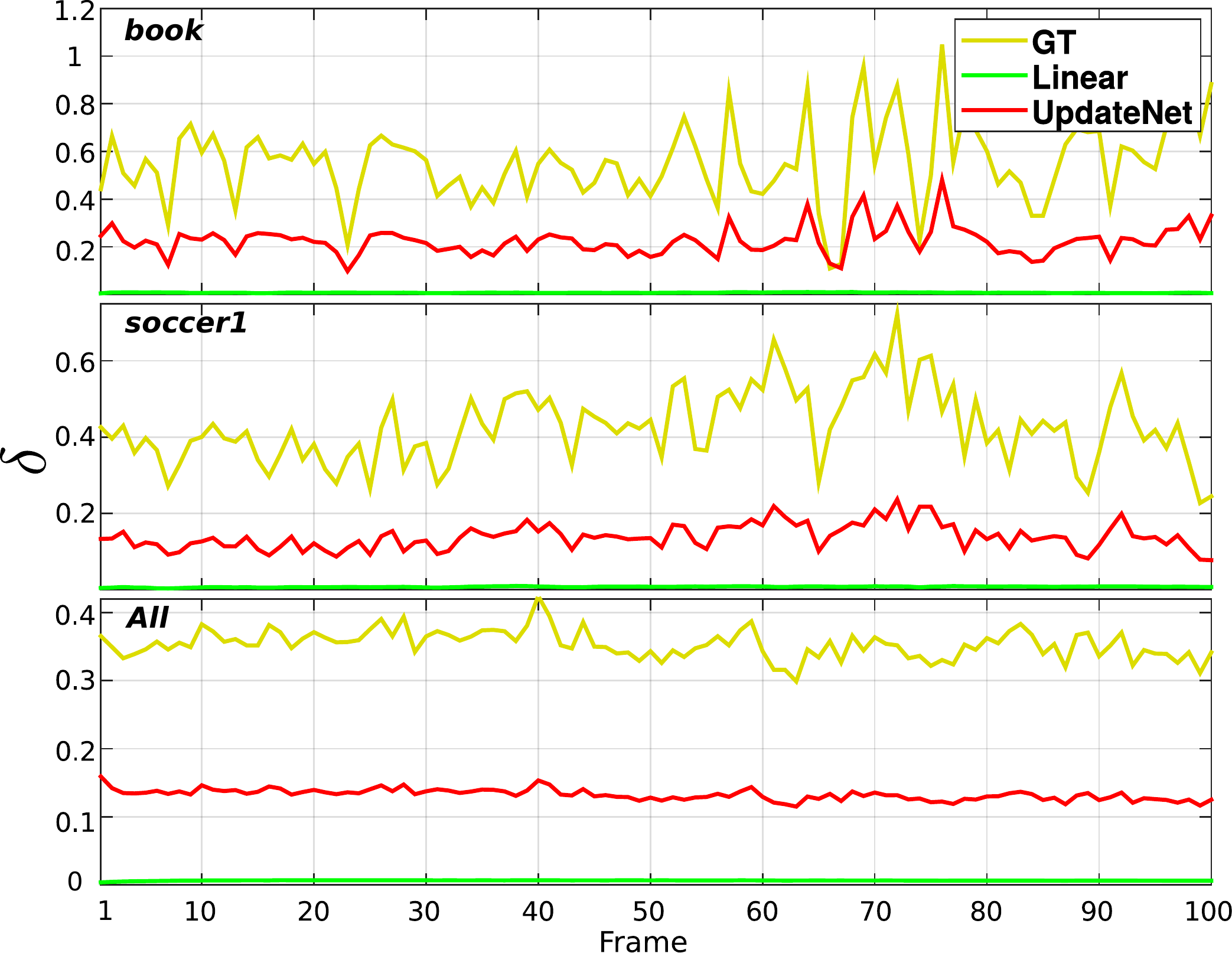}
    \caption{\small \textbf{Change rate between contiguous frames.} We present individual results for two example videos (top, middle) and average results for all videos in VOT2018.} 
    \vspace{-5mm}
    \label{fig:change_rate}
\end{figure}

In order to further study this observation, we propose quantifying the change rate between templates of contiguous frames.
For each $i\in \{1,...,N\}$ we compute the average difference in the template as $\delta_i = \frac{1}{|E|}\sum_{E} |T_i - T_{i-1}|$, where $N$ is the number of frames in a video and the sum runs over each element of the feature maps (e.g.\ $E=6\times6\times256$).
We present the results in Figure~\ref{fig:change_rate}.
The bottom row contains the average change rate $\delta$ of all 60 videos in VOT2018~\cite{kristan2018sixth}.
It is clear that the linear update strategy cannot deliver the updating rate required by the change in the features of the ground-truth template.
UpdateNet, on the other hand, provides a much more adaptive strategy that is substantially closer in magnitude to the change rate of the ground-truth template. 
The top and middle rows also show the change rate of the two individual sequences in Figure~\ref{fig:templates},  'book' and 'soccer1'.
We can see UpdateNet mimics the real template in high change periods, as indicated by the high correlation on their extremes.
This leads to predicting better response map as shown in Figure~\ref{fig:templates}.

\vspace{-1mm}
\subsection{Generality and tracking speed}
\begin{figure}[!t]
\centering
    \includegraphics[width=\columnwidth]{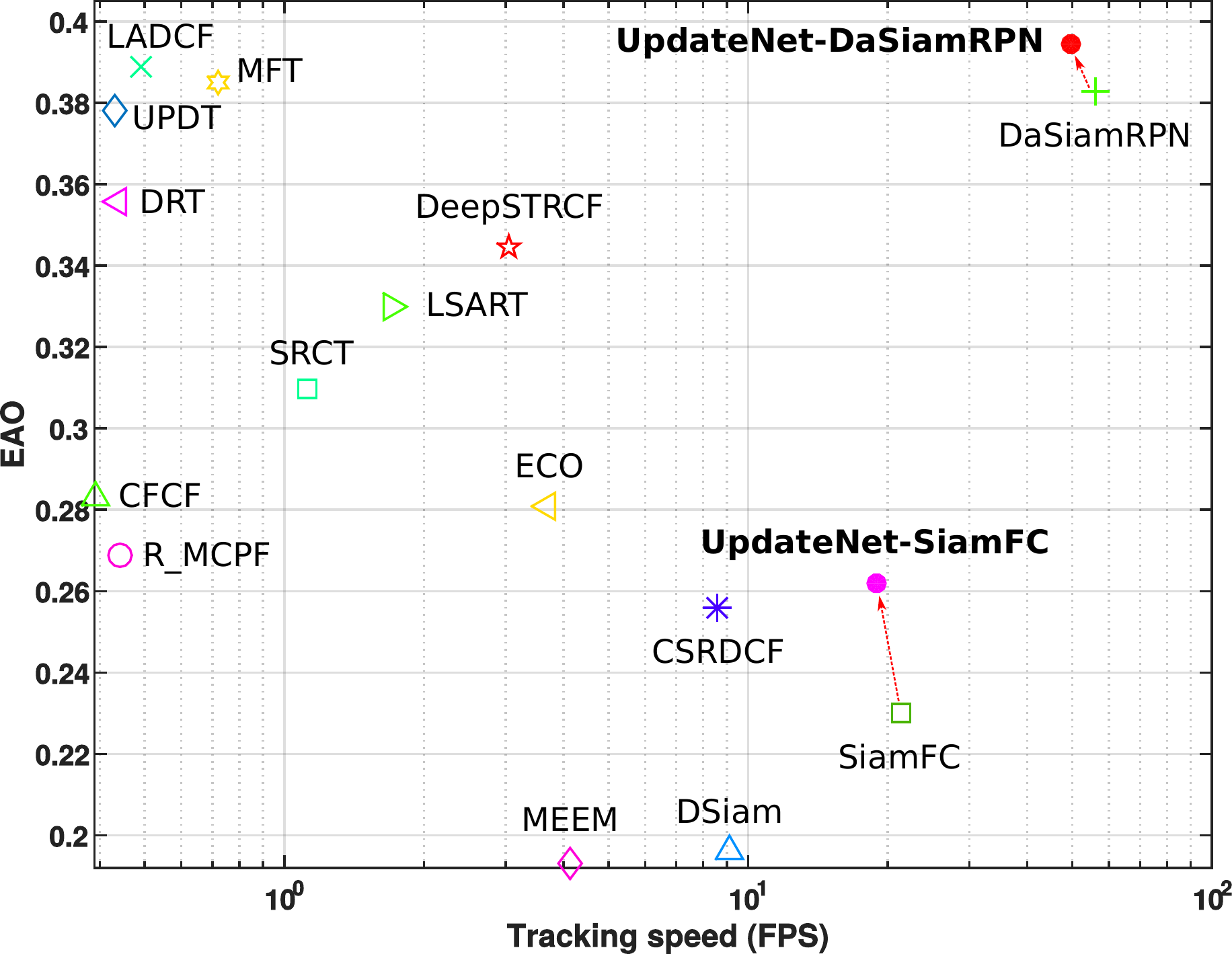}
    \caption{\textbf{EAO vs. speed on VOT2018.} We compare our UpdateNet combined with two different Siamese trackers against the state-of-the-art methods. UpdateNet can substantially improve the tracking performance without significantly affecting the speed.}
    \vspace{-2mm}    %
    \label{fig:eao_fps}
\end{figure}

\begin{figure}[t]
\centering
    \includegraphics[width=\columnwidth]{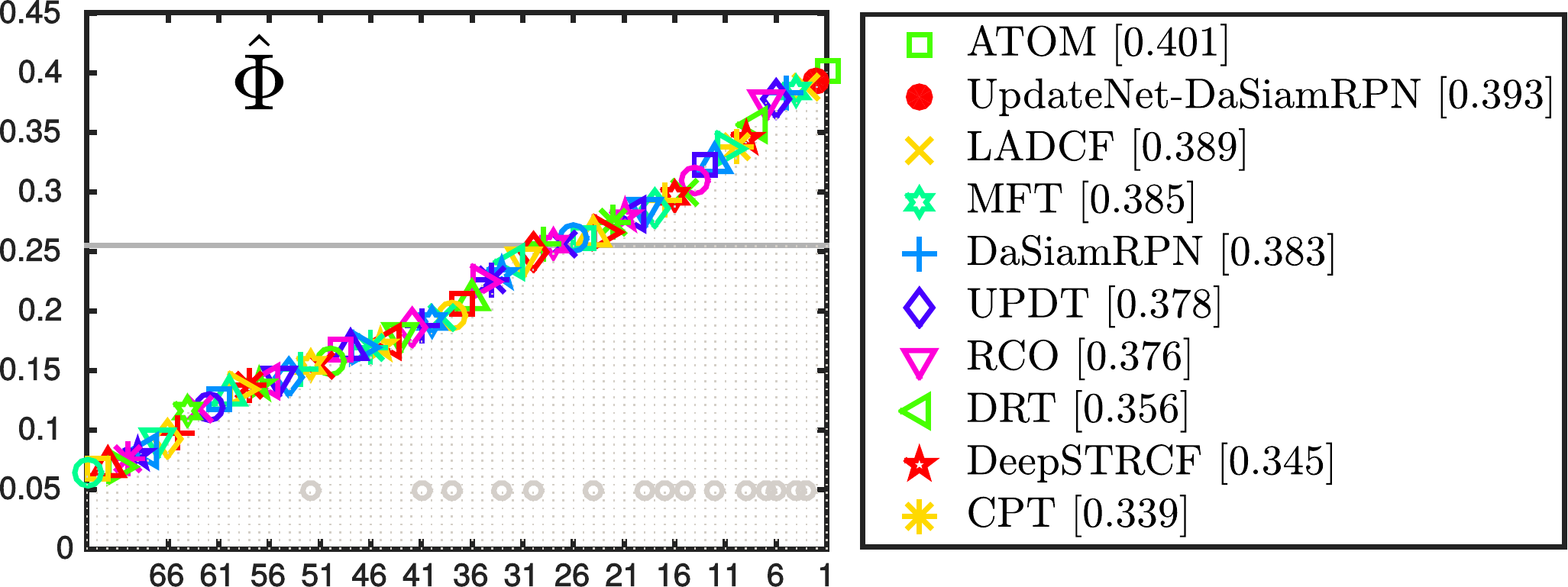}
    \caption{\textbf{EAO performance on VOT2018.} We compare our method with the state-of-the-art methods on VOT2018. Our proposed approach achieves superior performance.}%
    \vspace{-5mm}
    \label{fig:eao_rank_2018}
\end{figure}

\vspace{-1mm}
In this section, we evaluate the generality of our UpdateNet by applying it to other Siamese trackers as shown in Figure~\ref{fig:eao_fps}. It presents results on VOT2018 in terms of EAO with respect to the tracking speed.
We measure tracking speed in frames per second (FPS) and use a logarithmic scale on its axis.
We observe that we improve on Siamese trackers, e.g. SiamFC~\cite{bertinetto2016fully} and DaSiamRPN~\cite{zhu2018distractor} by adding a very small temporal overhead.
Finally the top-performance trackers are shown in Figure~\ref{fig:eao_rank_2018}.
We compare with trackers including DRT~\cite{sun2018correlation}, DeepSTRCF~\cite{li2018learning}, LSART~\cite{sun2018learning}, R\_MCPF~\cite{zhang2019learning}, SRCT~\cite{lee2018salient}, CSRDCF~\cite{lukezic2017discriminative}, LADCF~\cite{xu2018learning}, MFT~\cite{kristan2018sixth}, UPDT~\cite{bhat2018unveiling} and ATOM~\cite{danelljan2019atom} among others as in ~\cite{kristan2018sixth}. 
Among the top trackers, our approach achieves superior performance while maintaining a very high efficiency.
Further, our tracker obtains a performance relative gain of 2.8\% over the base tracker DaSiamRPN~\cite{zhu2018distractor}. 

\vspace{-1mm}
\subsection{Fine-tuning the linear update rate}
\begin{figure}[!t]
    \centering
    \includegraphics[width=.8\columnwidth]{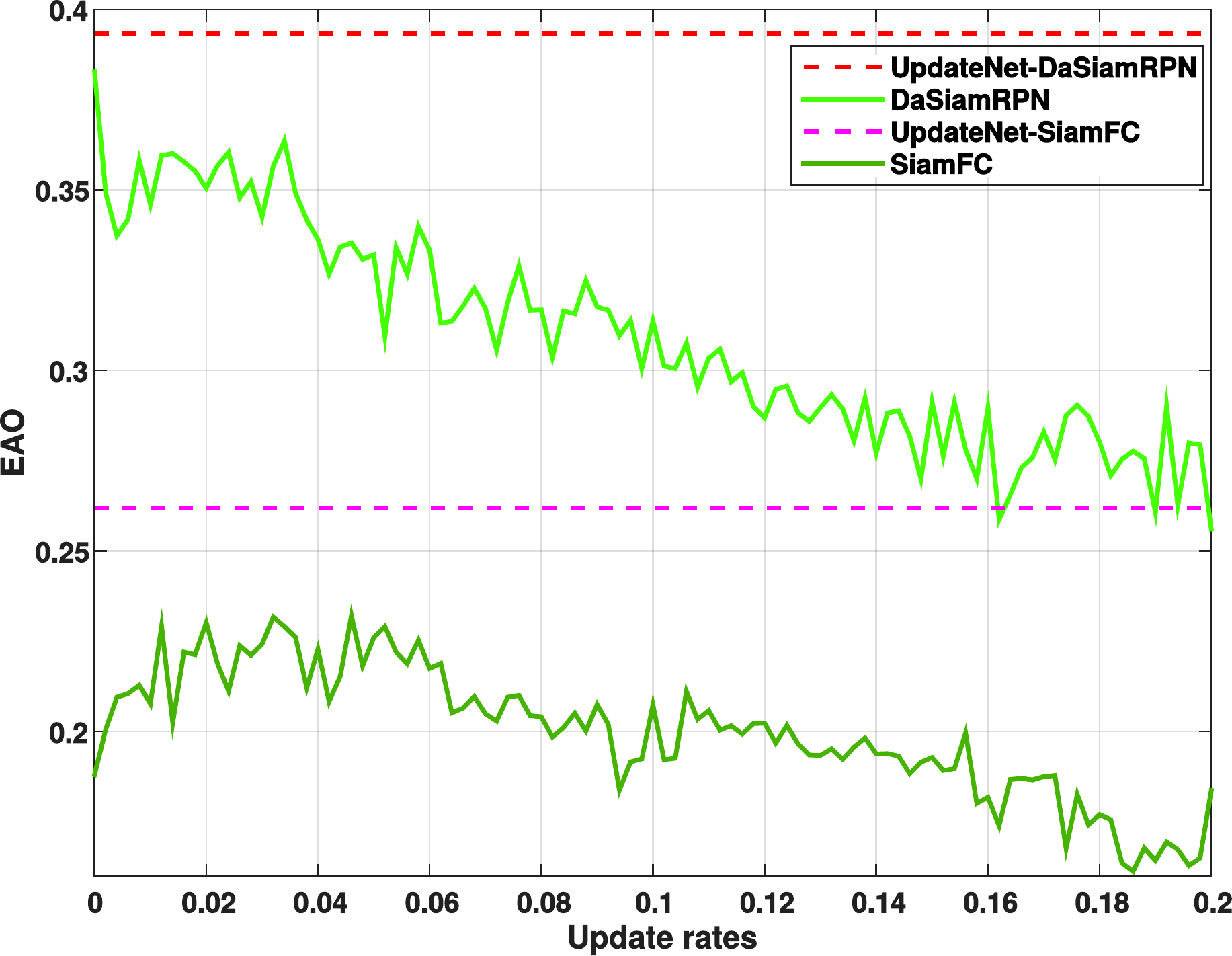}
     \caption{\textbf{The linear update rate evaluation for DaSiamRPN and SiamFC on VOT2018~\cite{kristan2018sixth}.} The x-axis is the update rate values. The y-axis is the EAO scores on VOT protocol~\cite{kristan2018sixth}. The red and pink dashed lines are our UpdateNet performances with DaSiamRPN and SiamFC, respectively.}
     \vspace{-1mm}
     \label{fig:linear_lrs_eao}
\end{figure}

\vspace{-1mm}
\label{sec:update_date}
The linear update in the previous section uses the update rate for SiamFC recommended by the authors~\cite{valmadre2017end} ($\gamma=0.0102$) and for DaSiamRPN from the original tracker~\cite{zhu2018distractor} ($\gamma=0$).
We now investigate whether the linear update strategy can bring higher performance gains when fine-tuning the update rate on the test set.
We test several update rates uniform sampled from the $[0,0.2]$ interval. 
Figure~\ref{fig:linear_lrs_eao} shows the performance of the linear update for DaSiamRPN (light green) and SiamFC (dark green). 
The red dashed line at the top and pink dashed line in the middle are the performance of our UpdateNet applied on DaSiamRPN and SiamFC respectively.
We can see for SiamFC the peak performance is indeed achieved between 0.01 and 0.05. 
For DaSiamRPN, the original tracker with no-update performs best, which proves that for more complex Siamese trackers trained off-line, on-line linear update may even damage the performance. 
This shows that even a fine-tuned linear update cannot improve its results any further.
Moreover, our UpdateNet outperforms all update rate values without requiring any manual fine-tuning. 
Despite the need of a higher update rate for some videos, we can see how the performance continuously and rapidly decreases as the update rate increases, evidencing the unsuitability of a fixed and general update rate for all videos.

\begin{table}[t]
    \begin{center}      
    \resizebox{\columnwidth}{!}{
        \begin{tabular}{ccccccc}
        \hline
 & DSiam & MemTrack & \multicolumn{2}{c}{SiamFC} & \multicolumn{2}{c}{DaSiamRPN}\\ 
 & \cite{guo2017learning} & \cite{yang2018learning} &  Linear & UpdateNet &  Linear & UpdateNet\\ \midrule
EAO &  0.181 & 0.273 & 0.235 & 0.289& \textcolor{blue}{\textbf{0.439}} & \textcolor{red}{\textbf{0.481}} \\ \hline
A &  0.492 & 0.533 & 0.529 & 0.543& \textcolor{red}{\textbf{0.619}} & \textcolor{blue}{\textbf{0.610}} \\ \hline
R  & 2.934 & 1.441 & 1.908 & 1.320& \textcolor{blue}{\textbf{0.262}} & \textcolor{red}{\textbf{0.206}}\\ \hline
        \end{tabular}}
        \vspace{0.5mm}
        \caption{\small \textbf{Results for other updating strategies on VOT2016.} DSiam~\cite{guo2017learning} and MemTrack~\cite{yang2018learning} use SiamFC as base tracker. The best two results are highlighted in red and blue fonts, respectively.}
        \vspace{-7mm}
        \label{table:vot2016}
    \end{center}
\end{table}

\vspace{-1mm}
\subsection{Comparison with other updating strategies} 
\vspace{-1mm}
Some recent approaches~\cite{guo2017learning,yang2018learning} proposed alternative updating strategies for Siamese trackers.
Table~\ref{table:vot2016} presents a comparison with DSiam~\cite{guo2017learning} and MemTrack~\cite{yang2018learning} on VOT2016, as~\cite{yang2018learning} only reports results on this VOT edition (see Figure~\ref{fig:eao_fps} for DSiam results on VOT2018).
Our UpdateNet leads to a more effective update and higher tracking performance, while also being applicable to different Siamese architectures. 
Despite the already excellent performance of DaSiamRPN, UpdateNet brings an improvement of 4.2\%, reaching state-of-the-art. 
Moreover, our approach yields a substantial absolute gain of 5.6\% in terms of robustness, which is a common weak point of Siamese trackers.

\vspace{-1mm}
\subsection{LaSOT dataset}
\begin{figure}[t!]
    \centering
    \includegraphics[width=\columnwidth]{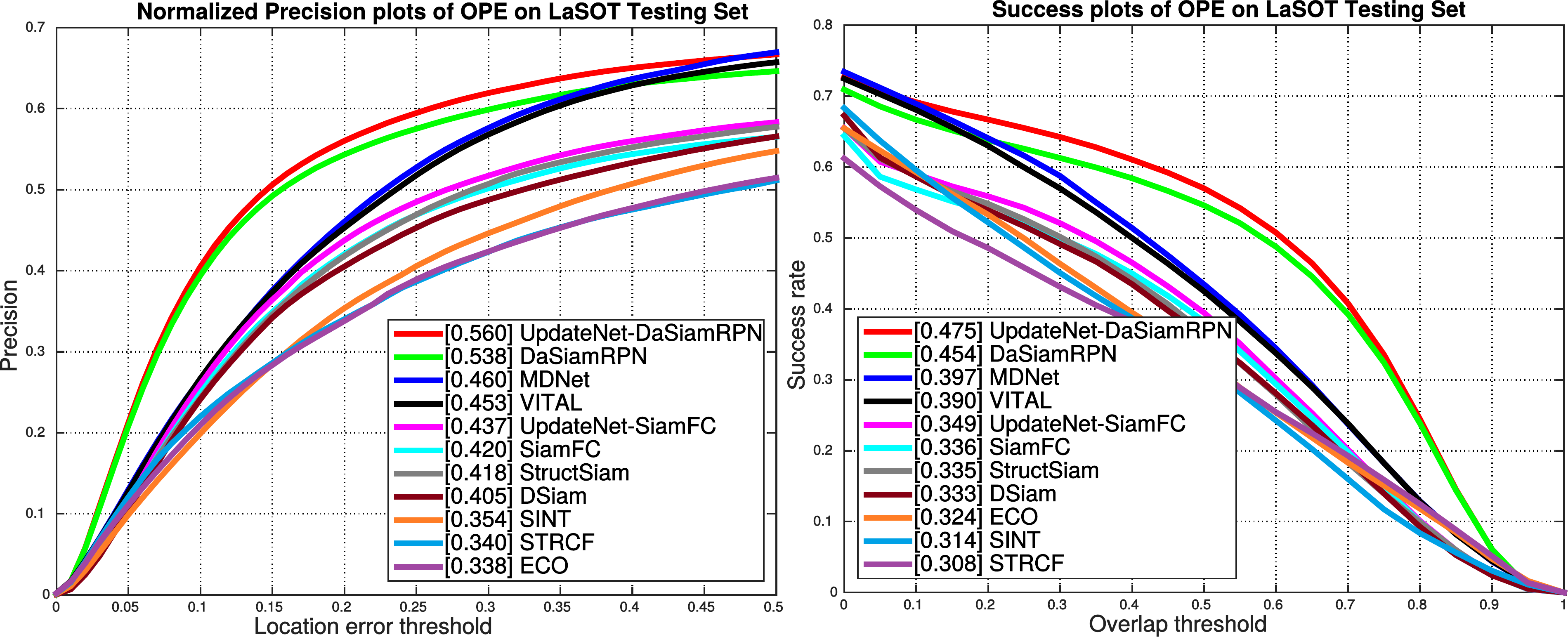}
     \caption{\textbf{Evaluation on LaSOT testing set.} Normalized precision and success plots following the OPE protocol II.} 
     \vspace{-1mm}
    \label{fig:lasot}
\end{figure}

\vspace{-1mm}
We test our model on the recent LaSOT dataset~\cite{fan2018lasot}.
Since long-term sequences are common in LaSOT, the updating component of the tracker is crucial, as more sudden variations may appear and object appearance may depart further from the input object template.
We show the top-10 trackers, including MDNet~\cite{nam2016learning}, VITAL~\cite{song2018vital}, StructSiam~\cite{zhang2018structured}, DSiam~\cite{guo2017learning}, SINT~\cite{tao2016siamese}, 
STRCF~\cite{li2018learning}, ECO~\cite{danelljan2017eco}, SiamFC~\cite{bertinetto2016fully} and DaSiamRPN~\cite{zhu2018distractor} in Figure~\ref{fig:lasot}.
The results are presented following the official protocol. 
We can see how UpdateNet enhances the updating capabilities of DaSiamRPN and leads to a significant performance boost on all measures.
As a result, our tracker with learned update strategy surpasses all state-of-the-art trackers on this dataset. 
This brings further evidence to the advantages of adaptive update strategy in terms of accurate object localization.

\vspace{-1mm}
\subsection{TrackingNet dataset}
\vspace{-1mm}
We evaluate our UpdateNet-DaSiamRPN on the testing set of TrackingNet~\cite{muller2018trackingnet} using their three evaluation metrics (Table~\ref{tab:trackingnet}).
Compared with DaSiamRPN, our UpdateNet+DaSiamRPN obtains absolute gains of 3.4\%, 1.9\% and 3.9\% in terms of precision, normalized precision and success. 
UpdateNet leads to a significant performance improvement on all three metrics.
This shows how a learning the model update can greatly benefit Siamese trackers on several datasets and under different measures.

\begin{table}[t]
    \begin{center}      
    \resizebox{1\columnwidth}{!}{
        \begin{tabular}{cccccccccc}
        \hline \vspace{1mm}
 & ATOM & ECO & CFNet & MDNet &  \multicolumn{2}{c}{SiamFC} &  \multicolumn{2}{c}{DaSiamRPN} \\ 
 & \cite{danelljan2019atom} & \cite{danelljan2017eco} &\cite{valmadre2017end} &\cite{nam2016learning}&Linear&UpdateNet & Linear & UpdateNet\\ \midrule
Precision (\%) & \textcolor{red}{\textbf{64.8}}  &49.2  &53.3 &56.5 &53.3&53.1 & 59.1  &\textcolor{blue}{\textbf{62.5}} \\ 
Norm. Prec. (\%) & \textcolor{red}{\textbf{77.1}}  &61.8  &65.4 &70.5&66.3 &67.4&73.3 &\textcolor{blue}{\textbf{75.2}}  \\ 
Success (\%)  &\textcolor{red}{\textbf{70.3}}  &55.4  &57.8 &60.6 &57.1 &58.4  &63.8   &\textcolor{blue}{\textbf{67.7}} \\ \hline
        \end{tabular}}
        \vspace{0.5mm}
        \caption{ \textbf{State-of-the-art comparison on TrackingNet.} Our UpdateNet significantly improves DaSiamRPN \cite{zhu2018distractor} with an absolute gain of 3.4\% and 3.9\%, in terms of precision and success. The best two results are highlighted in red and blue fonts, respectively.} 
        \label{tab:trackingnet}
        \vspace{-7mm}
    \end{center}
\end{table}

\vspace{-1.5mm}

\section{Conclusions}
\vspace{-1.5mm}

Siamese trackers usually update their appearance template using a simple linear update rule. 
We identify several shortcomings of this linear update and propose to learn the updating step as an optimization problem.
We employ a neural network, coined UpdateNet, that learns how to update the current accumulated template given the appearance template of the first frame, the current frame, and the accumulated template of the previous step.
The proposed UpdateNet is general and can be integrated into all Siamese trackers.
Comparable results on four benchmark datasets (VOT2016, VOT2018, LaSOT, and TrackingNet) show that the proposed approach to updating does significantly improve the performance of the trackers with respect to the standard linear update (or with respect to no-update at all). 

	\begin{small}
	\noindent\textbf{Acknowledgements.} We thank the Spanish project TIN2016-79717-R and mention	Generalitat de Catalunya CERCA Program. 
	\end{small}

{\small
\bibliographystyle{ieee}
\bibliography{shortstrings,egbib}

\begin{thebibliography}{10}\itemsep=-1pt

\bibitem{bertinetto2016fully}
L.~Bertinetto, J.~Valmadre, J.~F. Henriques, A.~Vedaldi, and P.~H. Torr.
\newblock Fully-convolutional siamese networks for object tracking.
\newblock In {\em ECCV workshop}, 2016.

\bibitem{bhat2018unveiling}
G.~Bhat, J.~Johnander, M.~Danelljan, F.~Shahbaz~Khan, and M.~Felsberg.
\newblock Unveiling the power of deep tracking.
\newblock In {\em ECCV}, 2018.

\bibitem{bolme2010visual}
D.~S. Bolme, J.~R. Beveridge, B.~A. Draper, and Y.~M. Lui.
\newblock Visual object tracking using adaptive correlation filters.
\newblock In {\em CVPR}, 2010.

\bibitem{bolme2009average}
D.~S. Bolme, B.~A. Draper, and J.~R. Beveridge.
\newblock Average of synthetic exact filters.
\newblock In {\em CVPR}, 2009.

\bibitem{choi2018context}
J.~Choi, H.~Jin~Chang, T.~Fischer, S.~Yun, K.~Lee, J.~Jeong, Y.~Demiris, and
  J.~Young~Choi.
\newblock Context-aware deep feature compression for high-speed visual
  tracking.
\newblock In {\em CVPR}, 2018.

\bibitem{choi2017visual}
J.~Choi, J.~Kwon, and K.~M. Lee.
\newblock Visual tracking by reinforced decision making.
\newblock {\em CoRR, abs/1702.06291}, 2017.

\bibitem{danelljan2017eco}
M.~Danelljan, G.~Bhat, F.~S. Khan, and M.~Felsberg.
\newblock Eco: efficient convolution operators for tracking.
\newblock In {\em CVPR}, 2017.

\bibitem{danelljan2019atom}
M.~Danelljan, G.~Bhat, F.~S. Khan, and M.~Felsberg.
\newblock Atom: Accurate tracking by overlap maximization.
\newblock In {\em CVPR}, 2019.

\bibitem{danelljan2015learning}
M.~Danelljan, G.~Hager, F.~Shahbaz~Khan, and M.~Felsberg.
\newblock Learning spatially regularized correlation filters for visual
  tracking.
\newblock In {\em ICCV}, 2015.

\bibitem{danelljan2016adaptive}
M.~Danelljan, G.~Hager, F.~Shahbaz~Khan, and M.~Felsberg.
\newblock Adaptive decontamination of the training set: A unified formulation
  for discriminative visual tracking.
\newblock In {\em CVPR}, 2016.

\bibitem{danelljan2016beyond}
M.~Danelljan, A.~Robinson, F.~S. Khan, and M.~Felsberg.
\newblock Beyond correlation filters: Learning continuous convolution operators
  for visual tracking.
\newblock In {\em ECCV}, 2016.

\bibitem{emami2012role}
A.~Emami, F.~Dadgostar, A.~Bigdeli, and B.~C. Lovell.
\newblock Role of spatiotemporal oriented energy features for robust visual
  tracking in video surveillance.
\newblock In {\em AVSS}, 2012.

\bibitem{fan2018lasot}
H.~Fan, L.~Lin, F.~Yang, P.~Chu, G.~Deng, S.~Yu, H.~Bai, Y.~Xu, C.~Liao, and
  H.~Ling.
\newblock Lasot: A high-quality benchmark for large-scale single object
  tracking.
\newblock {\em CoRR, abs/1809.07845}, 2018.

\bibitem{fan2018siamese}
H.~Fan and H.~Ling.
\newblock Siamese cascaded region proposal networks for real-time visual
  tracking.
\newblock {\em CoRR, abs/1812.06148}, 2018.

\bibitem{guo2017learning}
Q.~Guo, W.~Feng, C.~Zhou, R.~Huang, L.~Wan, and S.~Wang.
\newblock Learning dynamic siamese network for visual object tracking.
\newblock In {\em ICCV}, 2017.

\bibitem{he2018twofold}
A.~He, C.~Luo, X.~Tian, and W.~Zeng.
\newblock A twofold siamese network for real-time object tracking.
\newblock In {\em CVPR}, 2018.

\bibitem{he2016deep}
K.~He, X.~Zhang, S.~Ren, and J.~Sun.
\newblock Deep residual learning for image recognition.
\newblock In {\em CVPR}, 2016.

\bibitem{held2016learning}
D.~Held, S.~Thrun, and S.~Savarese.
\newblock Learning to track at 100 fps with deep regression networks.
\newblock In {\em ECCV}, 2016.

\bibitem{henriques2015high}
J.~F. Henriques, R.~Caseiro, P.~Martins, and J.~Batista.
\newblock High-speed tracking with kernelized correlation filters.
\newblock {\em TPAMI}, 37(3):583--596, 2015.

\bibitem{kiani2017learning}
H.~Kiani~Galoogahi, A.~Fagg, and S.~Lucey.
\newblock Learning background-aware correlation filters for visual tracking.
\newblock In {\em ICCV}, 2017.

\bibitem{kiani2015correlation}
H.~Kiani~Galoogahi, T.~Sim, and S.~Lucey.
\newblock Correlation filters with limited boundaries.
\newblock In {\em CVPR}, 2015.

\bibitem{kristan2018sixth}
M.~Kristan, A.~Leonardis, J.~Matas, M.~Felsberg, R.~Pflugfelder, L.~{\v{C}}.
  Zajc, T.~Voj{\'\i}r̃, G.~Bhat, A.~Luke{\v{z}}i{\v{c}}, A.~Eldesokey, et~al.
\newblock The sixth visual object tracking vot2018 challenge results.
\newblock In {\em ECCV workshop}, 2018.

\bibitem{VOT_TPAMI}
M.~Kristan, J.~Matas, A.~Leonardis, T.~Vojir, R.~Pflugfelder, G.~Fernandez,
  G.~Nebehay, F.~Porikli, and L.~\v{C}ehovin.
\newblock A novel performance evaluation methodology for single-target
  trackers.
\newblock {\em TPAMI}, 38(11):2137--2155, Nov 2016.

\bibitem{kristan2016VOT}
M.~Kristan, R.~Pflugfelder, A.~Leonardis, J.~Matas, F.~Porikli, L.~Cehovin,
  G.~Nebehay, G.~Fernandez, T.~Vojir, A.~Gatt, et~al.
\newblock The visual object tracking vot2016 challenge results.
\newblock In {\em ECCV workshop}, 2016.

\bibitem{lee2018salient}
H.~Lee and D.~Kim.
\newblock Salient region-based online object tracking.
\newblock In {\em WACV}, 2018.

\bibitem{li2018siamrpn++}
B.~Li, W.~Wu, Q.~Wang, F.~Zhang, J.~Xing, and J.~Yan.
\newblock Siamrpn++: Evolution of siamese visual tracking with very deep
  networks.
\newblock {\em CoRR, abs/1812.11703}, 2018.

\bibitem{li2018high}
B.~Li, J.~Yan, W.~Wu, Z.~Zhu, and X.~Hu.
\newblock High performance visual tracking with siamese region proposal
  network.
\newblock In {\em CVPR}, 2018.

\bibitem{li2018learning}
F.~Li, C.~Tian, W.~Zuo, L.~Zhang, and M.-H. Yang.
\newblock Learning spatial-temporal regularized correlation filters for visual
  tracking.
\newblock In {\em CVPR}, 2018.

\bibitem{liu2012hand}
L.~Liu, J.~Xing, H.~Ai, and X.~Ruan.
\newblock Hand posture recognition using finger geometric feature.
\newblock In {\em ICPR}, 2012.

\bibitem{lukezic2017discriminative}
A.~Lukezic, T.~Voj{\'\i}r, L.~C. Zajc, J.~Matas, and M.~Kristan.
\newblock Discriminative correlation filter with channel and spatial
  reliability.
\newblock In {\em CVPR}, 2017.

\bibitem{muller2018trackingnet}
M.~Muller, A.~Bibi, S.~Giancola, S.~Alsubaihi, and B.~Ghanem.
\newblock Trackingnet: A large-scale dataset and benchmark for object tracking
  in the wild.
\newblock In {\em ECCV}, 2018.

\bibitem{nam2016learning}
H.~Nam and B.~Han.
\newblock Learning multi-domain convolutional neural networks for visual
  tracking.
\newblock In {\em CVPR}, 2016.

\bibitem{park2018meta}
E.~Park and A.~C. Berg.
\newblock Meta-tracker: Fast and robust online adaptation for visual object
  trackers.
\newblock In {\em ECCV}, 2018.

\bibitem{renoust2016visual}
B.~Renoust, D.-D. Le, and S.~Satoh.
\newblock Visual analytics of political networks from face-tracking of news
  video.
\newblock {\em {IEEE} Transactions on Multimedia}, 18(11):2184--2195, 2016.

\bibitem{song2018vital}
Y.~Song, C.~Ma, X.~Wu, L.~Gong, L.~Bao, W.~Zuo, C.~Shen, R.~W. Lau, and M.-H.
  Yang.
\newblock Vital: Visual tracking via adversarial learning.
\newblock In {\em CVPR}, 2018.

\bibitem{stauffer1999adaptive}
C.~Stauffer and W.~E.~L. Grimson.
\newblock Adaptive background mixture models for real-time tracking.
\newblock In {\em CVPR}, 1999.

\bibitem{sun2018correlation}
C.~Sun, D.~Wang, H.~Lu, and M.-H. Yang.
\newblock Correlation tracking via joint discrimination and reliability
  learning.
\newblock In {\em CVPR}, 2018.

\bibitem{sun2018learning}
C.~Sun, D.~Wang, H.~Lu, and M.-H. Yang.
\newblock Learning spatial-aware regressions for visual tracking.
\newblock In {\em CVPR}, 2018.

\bibitem{tao2016siamese}
R.~Tao, E.~Gavves, and A.~W. Smeulders.
\newblock Siamese instance search for tracking.
\newblock In {\em CVPR}, 2016.

\bibitem{valmadre2017end}
J.~Valmadre, L.~Bertinetto, J.~Henriques, A.~Vedaldi, and P.~H. Torr.
\newblock End-to-end representation learning for correlation filter based
  tracking.
\newblock In {\em CVPR}, 2017.

\bibitem{wang2018learning}
Q.~Wang, Z.~Teng, J.~Xing, J.~Gao, W.~Hu, and S.~Maybank.
\newblock Learning attentions: residual attentional siamese network for high
  performance online visual tracking.
\newblock In {\em CVPR}, 2018.

\bibitem{wu2013otb}
Y.~Wu, J.~Lim, and M.-H. Yang.
\newblock Online object tracking: A benchmark.
\newblock In {\em CVPR}, 2013.

\bibitem{wu2015object}
Y.~Wu, J.~Lim, and M.-H. Yang.
\newblock Object tracking benchmark.
\newblock {\em TPAMI}, 37(9):1834--1848, 2015.

\bibitem{xu2018learning}
T.~Xu, Z.-H. Feng, X.-J. Wu, and J.~Kittler.
\newblock Learning adaptive discriminative correlation filters via temporal
  consistency preserving spatial feature selection for robust visual tracking.
\newblock {\em CoRR, abs/1807.11348}, 2018.

\bibitem{yang2018learning}
T.~Yang and A.~B. Chan.
\newblock Learning dynamic memory networks for object tracking.
\newblock In {\em ECCV}, 2018.

\bibitem{yao2018joint}
Y.~Yao, X.~Wu, L.~Zhang, S.~Shan, and W.~Zuo.
\newblock Joint representation and truncated inference learning for correlation
  filter based tracking.
\newblock In {\em ECCV}, 2018.

\bibitem{zhang2014fast}
K.~Zhang, L.~Zhang, Q.~Liu, D.~Zhang, and M.-H. Yang.
\newblock Fast visual tracking via dense spatio-temporal context learning.
\newblock In {\em ECCV}, 2014.

\bibitem{zhang2019learning}
T.~Zhang, C.~Xu, and M.-H. Yang.
\newblock Learning multi-task correlation particle filters for visual tracking.
\newblock {\em TPAMI}, 41(2):365--378, 2019.

\bibitem{zhang2018structured}
Y.~Zhang, L.~Wang, J.~Qi, D.~Wang, M.~Feng, and H.~Lu.
\newblock Structured siamese network for real-time visual tracking.
\newblock In {\em ECCV}, 2018.

\bibitem{zhu2018distractor}
Z.~Zhu, Q.~Wang, B.~Li, W.~Wu, J.~Yan, and W.~Hu.
\newblock Distractor-aware siamese networks for visual object tracking.
\newblock In {\em ECCV}, 2018.

\end{thebibliography}
}

\clearpage

\centerline{\textbf{\large{Supplementary Material}}}
\vspace{-4mm}
\beginsupplement

\section{Difference with offline weighted fusion}
\vspace{-2mm}
Another possible update mechanism that improves on the linear update could be learning an offline weighted fusion of three templates (the input of UpdateNet). In order to demonstrate the benefits of more sophisticated updating mechanisms, we propose the following experiment.

\begin{figure}[t!]
    \centering
    \vspace{-2mm}
    \includegraphics[width=.75\columnwidth]{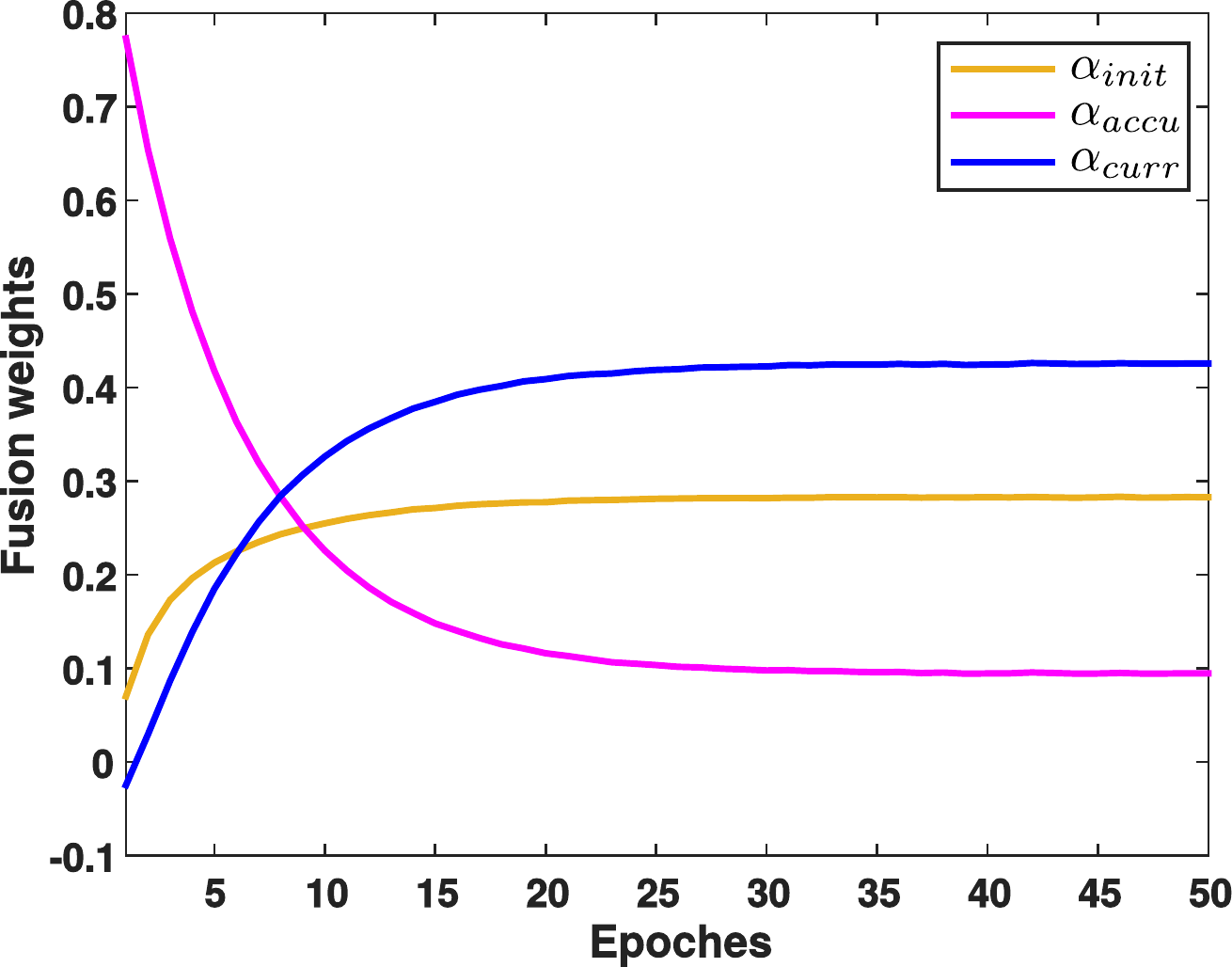}
     \caption{\small Training of the learned fusion weights offline.}
     \vspace{-7mm}
     \label{fig:alpha}
\end{figure}

UpdateNet uses a convolutional neural network that leverages previous templates to predict an accumulated template that is similar to the real one. Therefore, the fusion mechanism implemented by UpdateNet is more sophisticated than a simple offline weighted fusion, {which} depends on the actual input features and can be adapted accordingly. In order to confirm this, we propose here the following experiment to compare UpdateNet to an offline weighted fusion. We express the template update as a weighted linear combination $\widetilde{T}_{i}=\alpha_{init}T_0^{GT}+\alpha_{accu}\widetilde{T}_{i-1}+\alpha_{curr}T_i$. We initialize the three weights to 0, 0.9898, and 0.0102 respectively following the default settings for the linear update.
Then, we train these weights with the same training process as UpdateNet until convergence, as shown in Figure~\ref{fig:alpha}.

Next we compare the EAO results on VOT2018. SiamFC with an offline weighted fusion achieves \emph{0.198}, which is a little higher than the baseline linear update (\emph{0.188}) but much lower than \emph{0.262} achieved with {UpdateNet}. These results show that our {UpdateNet} is significantly better than an offline weighted fusion. 
We mainly attribute the superior results of UpdateNet to the following reasons. The offline weighted fusion learns a high value for  $\alpha_{curr}$. In this case, the tracker is likely to succumb to drift when the current template is not reliable.
Instead of excessively relying on the current template, UpdateNet can benefit from all the input templates due to the representation bottleneck in the channel dimension.
Furthermore, UpdateNet includes a non-linearity which allows it to better adjust to the non-linear variations, such as rotation and object motion.
Thus, yielding more expressive and reliable representations for the predicted template.

\vspace{-3mm}
\section{Visualization of updating templates}
\vspace{-2mm}

\begin{figure*}[hbt!]
    \centering
    \includegraphics[width=\textwidth]{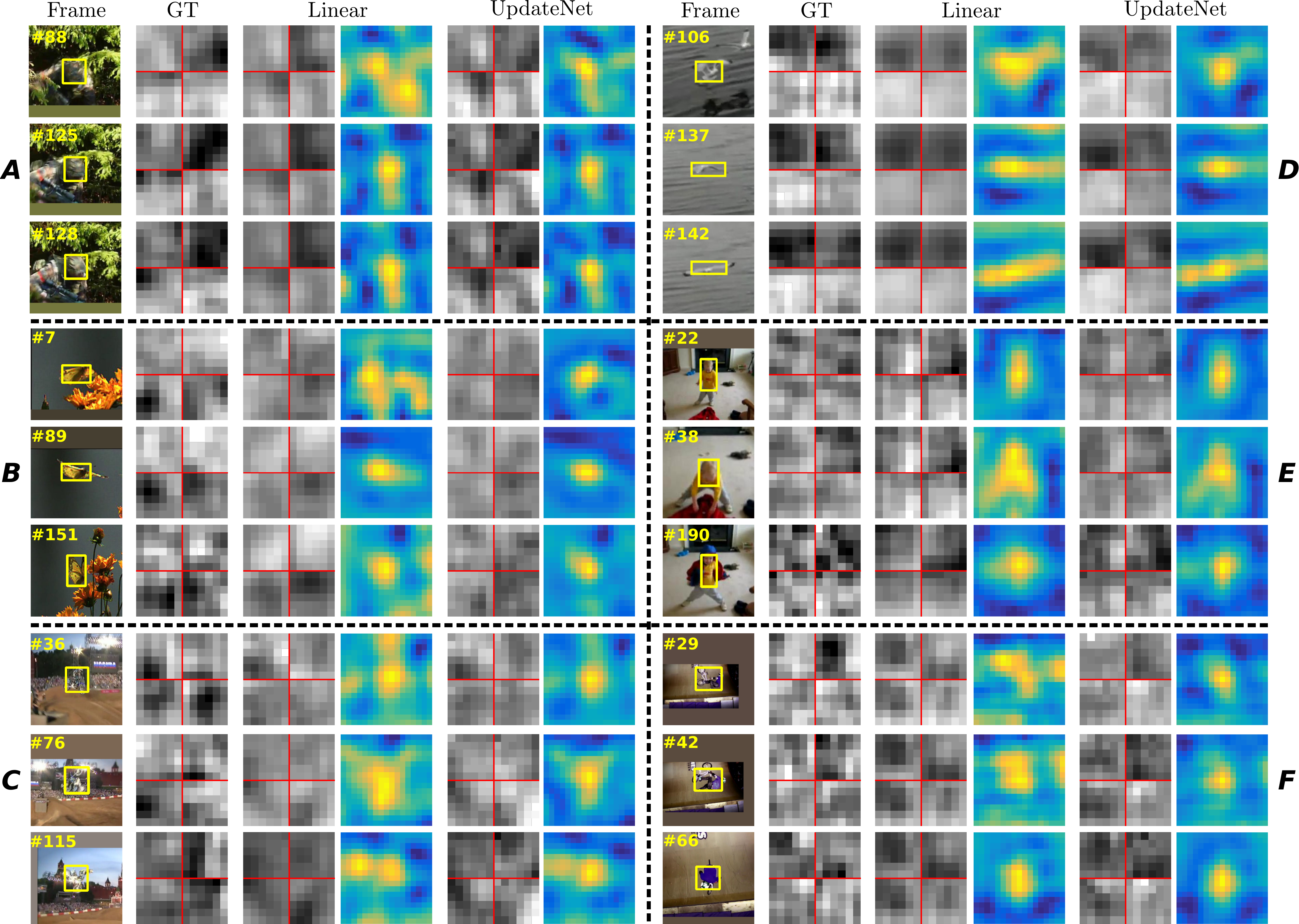}
     \caption{\small \textbf{Visualization accumulated and ground-truth templates for SiamFC.} The first column shows the search region and the ground-truth box. `GT' shows top four channels of the real template extracted from the ground-truth box. For each update strategy (`Linear' and `UpdateNet') we show the accumulated templates and the resulting response map when applied to the search region, respectively. }
     \label{fig:templates34}
\end{figure*}

We provide additional accumulated templates of SiamFC for both linear update and UpdateNet in Figure~\ref{fig:templates34} (similar to Figure 3 in the paper). By visualizing more exemplar videos, we can see that UpdateNet learns templates which are more similar to the ground-truth and predicts more accurate response maps for cross-correlation.
For visualization ease, we add a red cross `\textcolor{red}{\textbf{+}}' to split the four channels of the template feature.
The four channels are the most dynamic channels in the ground-truth template for the corresponding video.
We select them as follows.
For each $j\in \{1,...,C\}$ we compute the average difference in the template as $\Delta_j = \frac{1}{|N|}\sum_{N} \frac{1}{|A|}\sum_{A} |T_i^{GT} - T_{i-1}^{GT}|$, where $N$ is the number of frames in a video and the sum runs over the spatial area of each channel of the feature maps (e.g.\ $A=6\times6$). We select largest 4 channels in terms of $\Delta_j$.

We can observe multiple interesting behaviors in Figure~\ref{fig:templates34}. Firstly, the accumulated templates using UpdateNet resemble the ground-truth more closely than those with linear update, see e.g.\ the bottom-right channel (106) in frame 125 of \textit{\textbf{A}},  and the bottom-right channel (121) in frame 42 of \textit{\textbf{F}}, where the same highlighted region appears for both UpdateNet and the ground-truth. 
The template of the linear update, instead, does not resemble the ground truth for either of these two sequences.
In general, the accumulated templates from UpdateNet are almost as dynamic as the ground-truth templates, meaning that our UpdateNet can adapt to the template change in a video much better than linear update, which changes very slowly. 
Secondly, we can see in the cross-correlation response map how UpdateNet better predicts the object location, while linear update predicts many spurious peaks on the response map and the true peak in the center is less sharp, see e.g.\ frame 115 in example \textit{\textbf{C}} with an additional peak, frame 76 in example \textit{\textbf{C}} and frame 106 in example \textit{\textbf{D}} with blurred peaks, frame 88 in example \textit{\textbf{A}} and frame 7 in example \textit{\textbf{B}} with multiple peaks, among others. 
To summarize, Figure~\ref{fig:templates34} clearly shows that our strategy does not negatively interfere with the desired correlation properties of the learned features, on the contrary, it helps by adaptingly updating the templates.
On the other hand, the accumulated templates of the linear update change at a very slow rate and are inefficient in keeping up with the appearance variation exhibited in videos.

\vspace{-3mm}
\section{Change rate for update}
\begin{figure*}[hbt!]
    \centering
    \includegraphics[width=\textwidth]{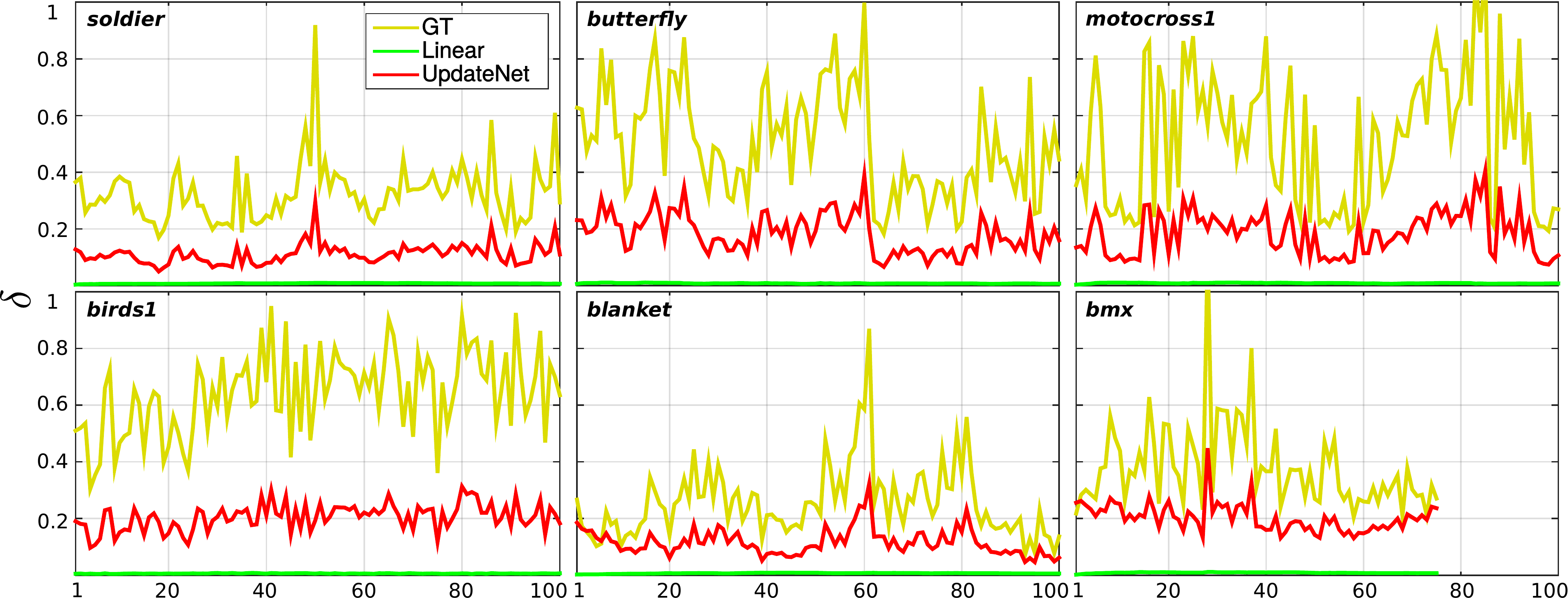}
    \caption{\small \textbf{Change rate between contiguous frames.} We present additional results for six example videos in VOT2018.}
    \label{fig:change_rate6}
\end{figure*}

\vspace{-2mm}
In addition to Figure 4 in the paper, we here provide similar results for the sequences shown in Figure~\ref{fig:templates34} of this supplementary material. 
We calculate the change rate $\delta$ between templates of contiguous frames and show the results in Figure~\ref{fig:change_rate6}. 
Our UpdateNet provides an adaptive update strategy that is close to the change rate of the real template, while linear update can only offer a constant change rate. 
The change rate for UpdateNet follows the same trends as the ground-truth, see for example the high correlation with the high peaks in e.g.\ frame 50 in `soldier', frame 60 in `butterfly' and frame 61 in `blanket'. This leads to predicting better response maps as shown in Figure~\ref{fig:templates34}.

\end{document}